\documentclass[runningheads]{llncs}

\usepackage{amsmath}
\usepackage{caption}
\usepackage{multirow}
\usepackage{xcolor}
\usepackage{sidecap}
\usepackage{hyperref}
\usepackage{adjustbox}
\usepackage{cite}
\usepackage{amsmath,amssymb,amsfonts}
\usepackage{algorithmic}
\usepackage{graphicx}
\usepackage{booktabs}
\usepackage[caption=false]{subfig}
\usepackage{layout}
\usepackage{color, colortbl}
\definecolor{LightCyan}{rgb}{0.88,1,1}
\usepackage{academicons}
\usepackage{pifont}
\usepackage{multirow}
\newcommand{\CC}{\cellcolor{LightCyan}}
\usepackage{textcomp}
\usepackage{adjustbox}
\usepackage{url}
\usepackage{soul}
\usepackage{float}

\makeatletter
\newcommand{\printfnsymbol}[1]{%
  \textsuperscript{\@fnsymbol{#1}}%
}
\newcommand{\etal}{\textit{et al.}}
\makeatother

\begin{document}

\title{WordVIS: A Color Worth A Thousand Words}
\titlerunning{WordVIS: A Color Worth A Thousand Words}
\author{Umar Khan* \and Saifullah, \and stefan agne \and andreas dengel \and Sheraz Ahmed}

\authorrunning{U. Khan et al.}
%\institute{
%	German Research Center for Artificial Intelligence (DFKI)
%	67663 Kaiserslautern\\\email{\{firstname.lastname\}@dfki.de}\\ \and
%	RPTU Kaiserslautern-Landau, 67663 Kaiserslautern, Germany\and
%	DeepReader GmbH, 67663 Kaiserlautern, Germany\\
%}
\institute{Deutsches Forschungszentrum für Künstliche Intelligenz, DFKI Kaiserrslautern, Germany.}

\maketitle
\begin{abstract}
Document classification is considered a critical element in automated document processing systems. In recent years multi-modal approaches have become increasingly popular for document classification. Despite their improvements, these approaches are underutilized in the industry due to their requirement for a tremendous volume of training data and extensive computational power. In this paper, we attempt to address these issues by embedding textual features directly into the visual space, allowing lightweight image-based classifiers to achieve state-of-the-art results using small-scale datasets in document classification. To evaluate the efficacy of the visual features generated from our approach on limited data, we tested on the standard dataset Tobacco-3482. Our experiments show a tremendous improvement in image-based classifiers, achieving an improvement of $4.64$\% using ResNet50 with no document pre-training. It also sets a new record for the best accuracy of the Tobacco-3482 dataset with a score of $91.14$\% using the image-based DocXClassifier with no document pre-training. The simplicity of the approach, its resource requirements, and subsequent results provide a good prospect for its use in industrial use cases.
\end{abstract}

\keywords{ document image classification \and document classification \and image processing \and data pre-processing \and visual embeddings \and textual embeddings \and deep learning.}

\section{Introduction}
\label{sec:introduction}
Documents play an integral role in modern business communication and record keeping. As a result, there is a growing interest today in the development of automated document processing pipelines in business workflows \cite{doc-img-cls-deep-cnns-Afzal2017,vibertgrid,doc-mm-cls-layoutlmv2-Xu2021,pick}. Document classification is a common starting point for many of these pipelines. Early classification of documents not only facilitates filtering, searching, and retrieval, but also enhances downstream performance \cite{doc-img-cls-deep-cnns-Afzal2017,doc-img-cls-first-cnn-Kang2014,doc-img-cls-stacked-cnns-Das2018}. For example, if it is possible to categorize a specific class with a high degree of confidence, an efficient information extraction module may be developed particularly for that class, thereby enhancing the pipeline's overall efficiency. Due to its fundamental importance, document classification has been extensively studied over the past few decades \cite{doc-cls-trad-Francesca2001,doc-cls-trad-structural-similarity-Dengel1995,doc-cls-trad-feature-sim-forest-kumar2014,doc-img-cls-deep-cnns-Afzal2017,doc-mm-cls-layoutlmv2-Xu2021} and has been widely adopted by the industry.	

Deep learning has been extensively studied in recent years in the context of document classification, resulting in a variety of approaches both in the image domain and in the multimodal domain. As a result, the task of building a high-performance document classifier is no longer considered arduous. As long as sufficient computing resources and training data are available, there are numerous state-of-the-art approaches \cite{doc-img-cls-effnet-Ferrando2020,doc-img-cls-docxclassifier-saifullah2022,doc-mm-cls-layoutlmv2-Xu2021,doc-mm-cls-Powalski2021,doc-mm-cls-two-stream-Asim2019} that can be directly applied to both large and small datasets to produce exceptional results in document classification. Especially interesting are the recent multimodal approaches which use both visual and textual features to perform the classification task and are particularly successful at countering the problem of high inter-class similarity and intra-class variance commonly found in documents \cite{doc-img-cls-rvlcdip-Harley2015,doc-img-cls-first-cnn-Kang2014}. However, there are several challenges involved with these approaches from a deployment perspective. Firstly, many of these approaches involve multiple streams of networks \cite{doc-mm-cls-two-stream-Asim2019,doc-mm-cls-two-stream-Noce2016} or large multimodal transformer networks \cite{doc-mm-cls-layoutlmv2-Xu2021,doc-mm-cls-Powalski2021}, which greatly increase the computational load. Additionally, such models are particularly challenging to train, since they require self-supervised learning across millions of data points \cite{doc-mm-cls-layoutlmv2-Xu2021,doc-mm-cls-layoutlm-Xu2020}. This can be especially problematic for small businesses, which have limited computing resources or training data, making it difficult to deploy such approaches in a practical manner. In addition to these issues, most existing multimodal approaches require feeding the textual and layout information directly to the model \cite{doc-mm-cls-layoutlmv2-Xu2021,doc-mm-cls-layoutlm-Xu2020,doc-mm-cls-two-stream-Asim2019}, which may require overhauling an existing document processing system. %Additionally, multi-modal techniques are not language agnostic and require additional training data. 
Finally, such multimodal techniques can also be difficult to extend to new languages and require additional training data for each target language.

% this was to be merged down
In this paper, we attempt to counter the aforementioned issues in multimodal approaches and present a lightweight approach for document classification that utilizes both visual and textual features of a document image without the need for any kind of self-supervised pretraining. Contrary to most existing multi-modal approaches, we embed the textual semantic features and context directly into the visual space of a document by assigning an RGB color to each word in accordance with the similarity of different letters.
% Fig.~\ref{fig:pipeline} provides an example of a color-coded document image resulting from our approach. As shown, our approach simply acts as a preprocessing step to a convert a plain document image into a color-encoded document image.
% this was to be merged up
% So in our publication, we have tried to come up with a novel text-to-visual space embedding technique inspired by the string metric concept of Levenshtein distance from Information Theory. Our method allows us to build a generalized method for embedding textual features as well as allows us to embed dataset-spe- cific textual features without requiring a change in the underlying mechanism of the method. 
% In this manner, our approach not only allows existing image-based classifiers to directly exploit the textual cues of a document but also considerably reduces the performance overhead associated with the processing of multi-modal data. 
% In addition, it requires no additional pretraining to learn the textual embeddings as in the case of typical multi-modal approaches, it is particularly suitable for small datasets. 
% In addition, our approach relies on text semantics and is therefore directly applicable to any language without the need for additional language-specific data, as is the case with transformer-based approaches. Finally, due to the simplicity of our approach, it can easily be integrated into existing CNN-based document classification pipelines.
In this manner, our approach not only allows existing image-based classifiers to directly exploit the textual cues of a document but also substantially reduces the performance overhead associated with the processing of multi-modal data. In addition, it requires no additional pretraining to learn the textual embeddings as in the case of typical multi-modal approaches, making it particularly suitable for small datasets. Due to the simplicity of our approach, not only can it easily be integrated into existing CNN-based document classification pipelines but can also be directly extended for any new languages without the need for additional language-specific data, as is the case with most transformer-based approaches.
The overall contributions of our paper can be summarized as follows:
\begin{itemize}
    \item[--] We present a novel approach for embedding textual semantic and contextual features in the visual space of a document. This enables training high-performing document classifiers in data-scarce settings. 
    \item[--] For the ablation study, we evaluate our approach with multiple CNN-based architectures on a small-scale document benchmark dataset Tobacco-3482 and show that our approach results in consistent performance improvements ranging from $3-5\%$ simply by training the models on the dataset.
    \item[--] WordVis was also able to improve the performance of the state-of-the-art DocXClassifier-B, ultimately resulting in a new best-record accuracy score of $91.14\%$ without the use of RVL-CDIP document pre-training, which means enabling the development of sustainable classifiers without extensive training, especially in data-scarce situations.
    
\end{itemize}

\section{Related Work}
\subsection{Document Image Classification}
The field of document image classification has evolved considerably over the past few decades. The early works in the field were primarily based on structural similarities between documents \cite{doc-cls-trad-structural-similarity-shin2016}, feature matching \cite{doc-cls-trad-feature-sim-forest-kumar2014}, or their combination \cite{doc-cls-trad-feature-based-structural-sim-Thompson2002}. In a different direction, classical machine learning-based approaches such as K-Nearest Neighbors (KNN) \cite{doc-cls-trad-knn-Baldi2003}, Decision Trees \cite{doc-cls-trad-Francesca2001} and Hidden Markov Models (HMM) \cite{doc-cls-trad-hmm-Diligenti2003} have also been explored for this task. A detailed overview of these approaches can be found in the survey paper by Chen \etal \cite{doc-cls-trad-survey-Chen2007}.
% Document classification is a legacy task in the document information extraction pipelines. And the way we have handled this task has tremendously evolved. Earlier works in the field highly based on developing heuristics for the task such as Chen et al [9], shin et al. [10], and Appiani et al [11]. A very well-structured summary of all these heuristics-based methods along with some early machine-learning-based approaches until 2006 is given in the survey paper by Chen et al [12]. With the advancements in hardware deemed crucial for deep learning and improvements in open-source machine learning libraries for deep learning, we saw the adoption of deep learning for such legacy tasks in document information extraction pipelines.

% With the seminal work by Schmidhuber \etal \cite{schmidhuber} in 2014 presenting one of the pioneering papers in introducing CNNs alongside LeCun et al \cite{lecun15} in 2015, It didn't take long before these were adopted for use in document classification. 
It was not long after the seminal work by Krizhevsky \etal \cite{nips_imagenetcnn_12} in which the popular AlexNet architecture was introduced for natural image classification, that a range of deep-learning based document classification systems were introduced. 
Kang \etal \cite{doc-img-cls-first-cnn-Kang2014} were the first to demonstrate the effectiveness of a neural network for document classification, which was significantly more successful compared to classical approaches. 
Later, Afzal investigated the use of much deeper CNNs for the classification of documents, as well as the advantages of transfer learning for document classification. Since then, this trend has continued steadily with the use of newer versions and variations of CNNs for the classification of documents \cite{doc-img-cls-effnet-Ferrando2020,doc-img-cls-docxclassifier-saifullah2022}. As CNN-based approaches have started to reach diminishing returns for this task, there has also been a growing interest in multi-modal approaches that combine the image, layout, and textual information of the document to perform the classification task. These techniques are generally implemented either in a multi-stream fashion which uses separate streams for visual, textual or other layout features \cite{doc-mm-cls-two-stream-Asim2019,doc-mm-cls-two-stream-Noce2016} or based on multimodal transformer architectures \cite{doc-mm-cls-layoutlmv2-Xu2021,doc-mm-cls-layoutlm-Xu2020} that are trained in a self-supervised manner on millions of training samples.
% Starting with Kang et al [15] explore the use of these networks in document classification. Whereas, a more visual-feature-focused work from Afzal et al [16] also explored the transfer learning aspect of CNNs for the use case of document classification. The continuation of this trend has since been at a steady pace with the use of newer versions and variations of CNNs being opted for the use case of document classification. However, with the use of multi-modal classifiers more recently. An example of this case is the LayoutLM by Xu et al [19] using visual, textual and layout features all in combination in transformer architecture has drastically improved results and has started to outperform the conventional CovNets. With these improvements, the focus has shifted towards forming much larger architectures making use of multiple-modalities. This approach has its limitations with massive computer power and training data requirements for pre-training these architecture, which is not a bigger problem for most larger players in the industry, however poses a much larger challenge for smaller companies and organizations alike.

\subsection{Visual Encoders for Textual Information}
In this section, we review some previous works for encoding textual semantics or contextual information into visual space in document analysis. Anoop \etal \cite{chargrid} proposed Chargrid, a grid-based color coding scheme for document images where the bounding box region of each character is colorized on a separate image mask based on a predetermined encoding scheme. The original image and its corresponding colorized image mask were then used to train a neural network for the task of key information extraction (KIE). To introduce more contextual and semantic information into the textual encoding, Timo \etal \cite{bertgrid}, introduced BERTGrid for the KIE task in which instead of colorizing the document in RGB space, instead they concatenate the BERT-based \cite{Devlin2019-bert} textual embeddings with the image. In particular, for each word, its corresponding BERT embedding is concatenated on top of the image within its bounding box. While this approach provided richer contextual information in comparison to Chargrid, it came with huge computational costs and increased dimensionality of the image input. Lin \etal \cite{vibertgrid} presented a similar approach to BERTGrid for KIE task but instead introduced the concatenation step of textual embedding into visual space at an intermediate feature map of a Convolutional Neural Network (CNN) instead of concatenating it with the input image. In a slightly different direction, Saman \etal \cite{saman_dicta18} used a simple color encoding scheme with the idea of distinguishing between numbers and alphabets to colorize the document images for the task of table detection in document images. As mentioned above, while some work has been done to encode textual information into visual space, it has not been explored in the context of document classification.

\begin{figure}[t]
    \centering   
        \includegraphics[width=\textwidth]{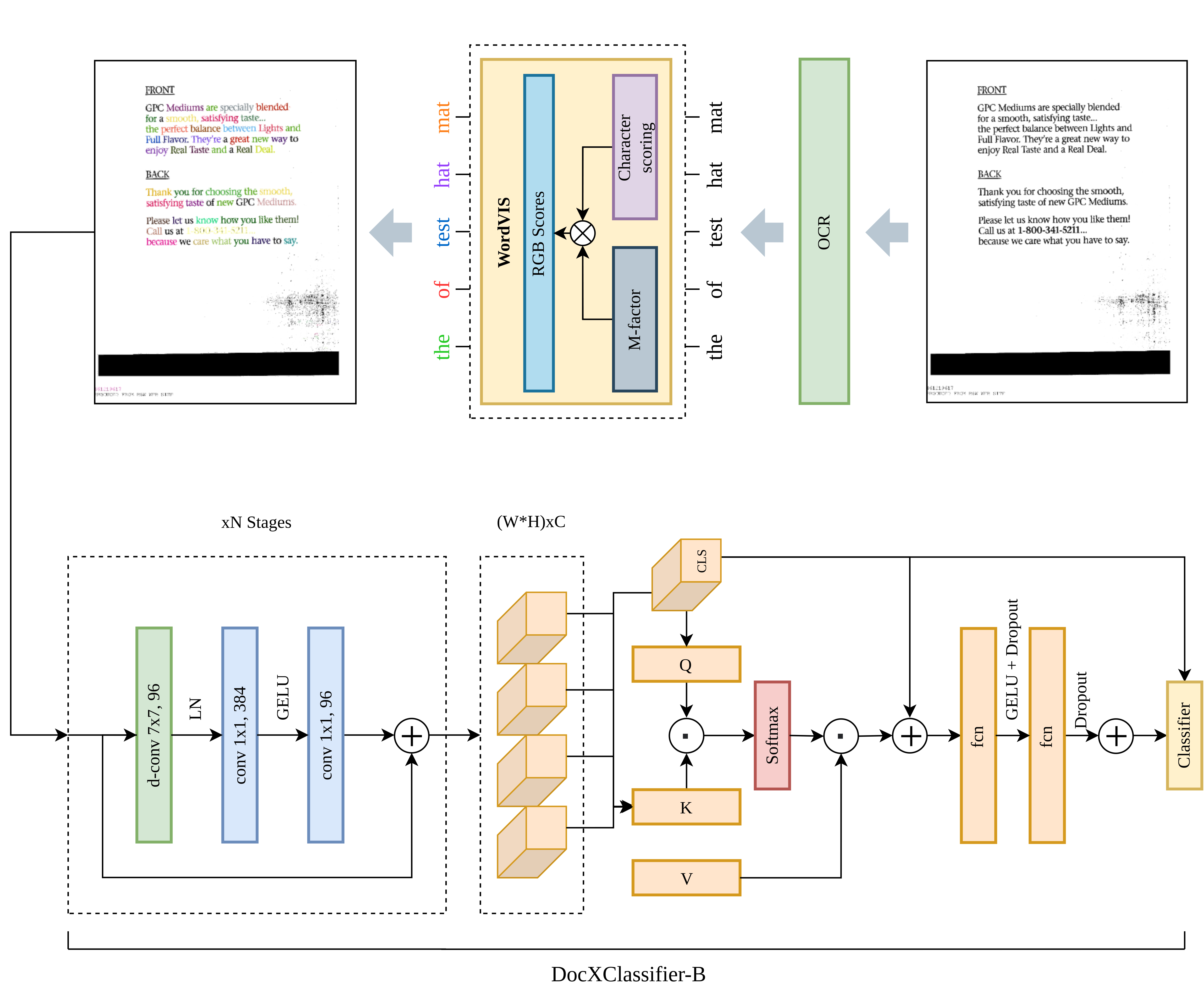}
    \caption{WordVIS as a pre-processing step for existing document classification models. Input document images are first passed through an OCR system to extract textual information. The textual features are then encoded within the visual space of the images using our approach. Finally, the pre-processed images are fed into a document classification model. Shown here is the DocXClassifier-B model.}
    \label{fig:pipeline}
\end{figure}
% Overall, our proposed approach is similar to that by Saman \etal \cite{saman_dicta18} but instead we colorize each word words based on language semantics. 

% Some work has already been done in order to bring ConvNets head to head in performance to these data-hungry architectures by hand-crafting features. Such as the application of distance transform by Gilani et al [20] on textual data for table detection. Similarly [21] made use of color encoding to colorize different regions based on semantic understanding. However, we feel in terms of document classification we see a lack of a generalized method for embedding textual information inside the visual space. Such a method would allow us to embed textual context information into the visual space which would greatly help the network extract information. This has been verified in different document tasks by Shahzad et al [22], where they have successfully shown that albeit Deep Learning great at learning representation greatly benefit from handcrafted features. Following with this would not only reduce the data overhead required for pre-training backbones as most image-based classifiers are pre-trained on ImageNet, but would also allow us to reduce the compute overhead by using a single modality for the classification of documents.

\section{WordVis: The Proposed Approach}

WordVIS is a novel pre-processing method that generates enhanced document representations using OCR data.
It generates enriched visual representation by encoding text to colors using a score lookup table and applies color masks on words using the original document image. The scoring mechanism was inspired by the fundamental concept of string metric calculation from information theory called "Levenshtein Distance". We leverage the core concept of how the distance is measured in strings. In this section, we describe the process of generating these document representations using WordVIS.

\begin{figure}[H]
\centering
\captionsetup[subfigure]{labelformat=empty}
\captionsetup[subfloat]{farskip=2pt,captionskip=2pt,singlelinecheck=false,justification=centering}
\subfloat{\includegraphics[width=0.40\linewidth]{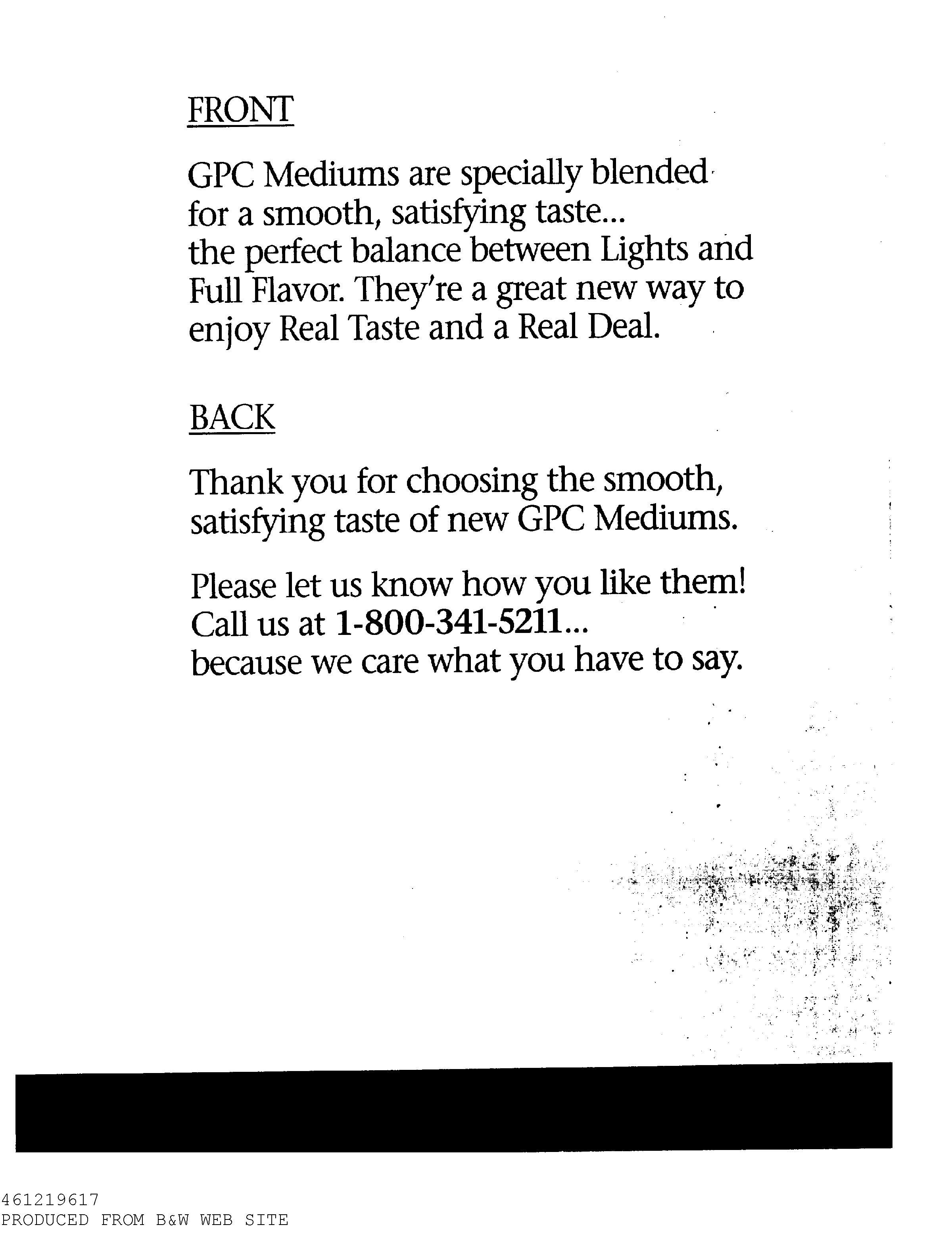}}\hfill
\subfloat{\includegraphics[width=0.40\linewidth]{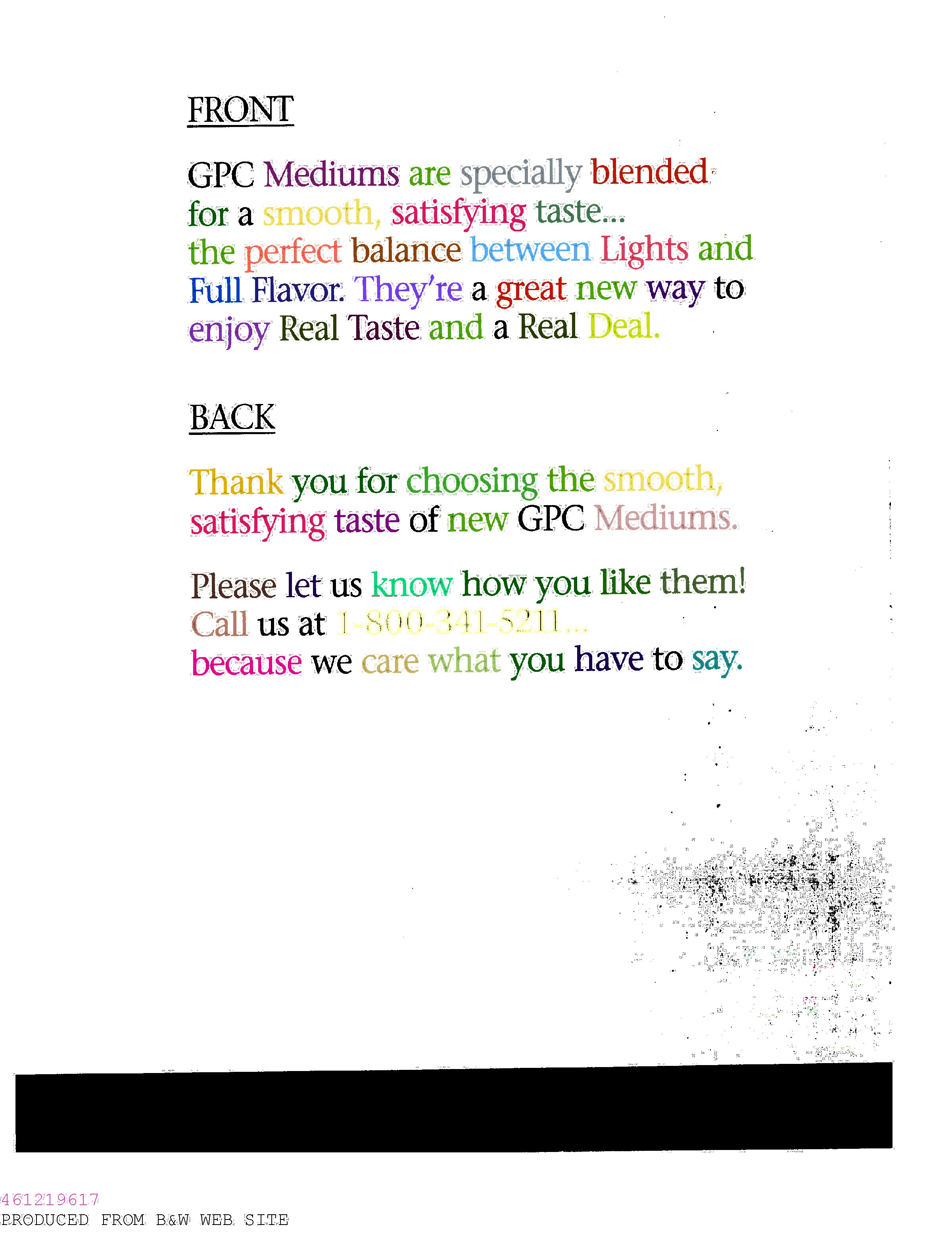}}\hfill
\\
\subfloat{\includegraphics[width=0.40\linewidth]{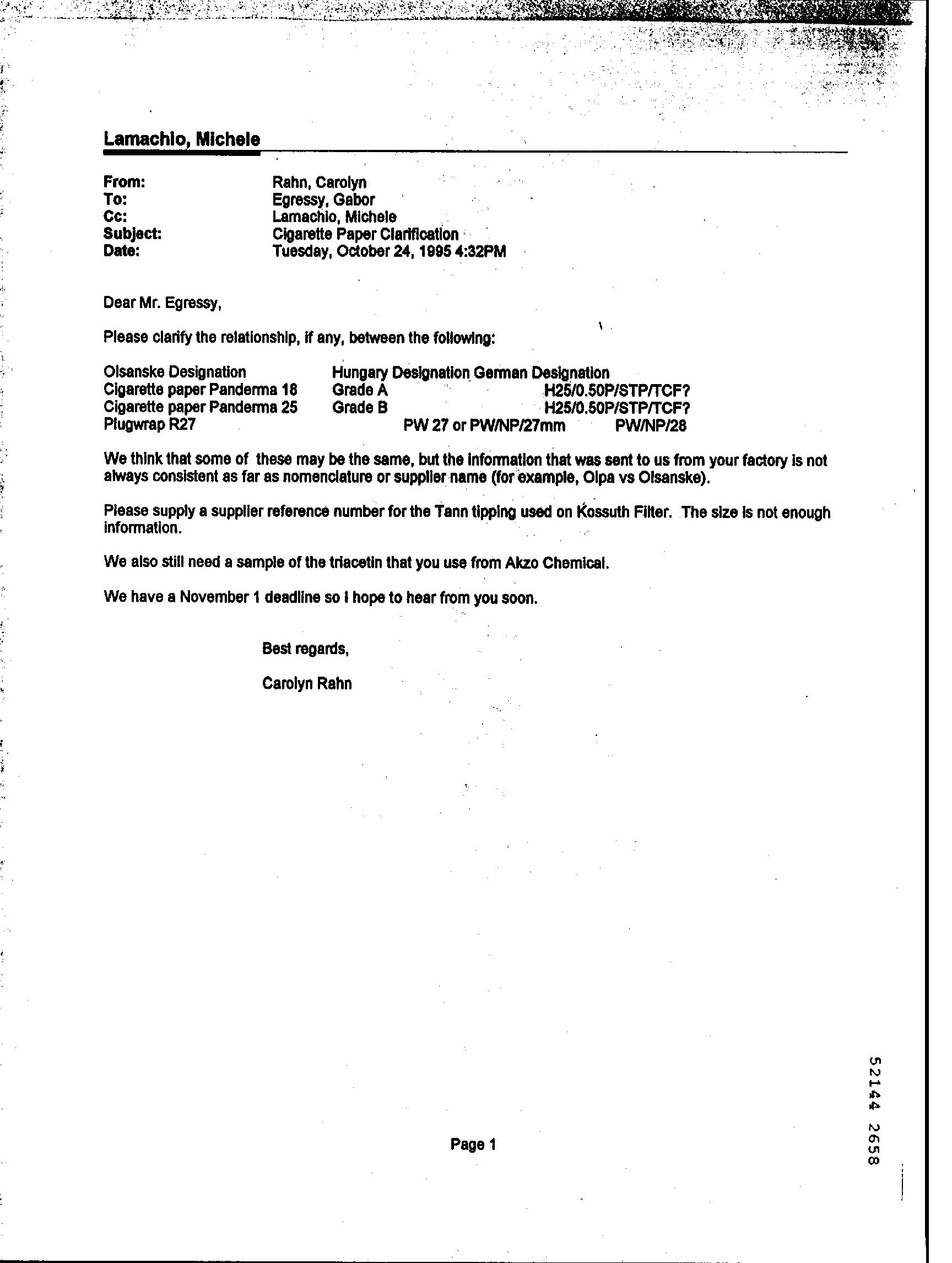}}\hfill
\subfloat{\includegraphics[width=0.40\linewidth]{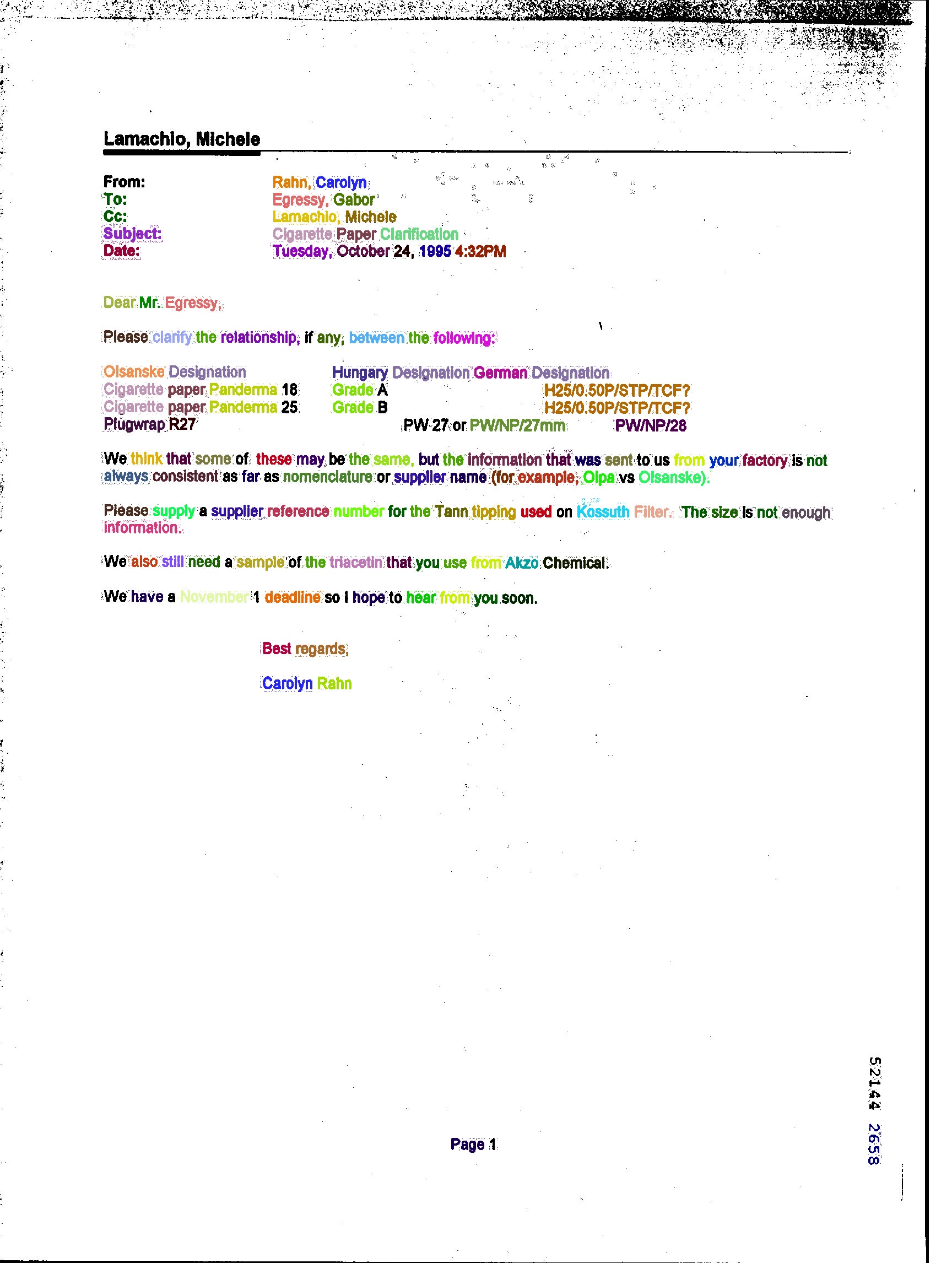}}\hfill
\\
\subfloat{\includegraphics[width=0.40\linewidth]{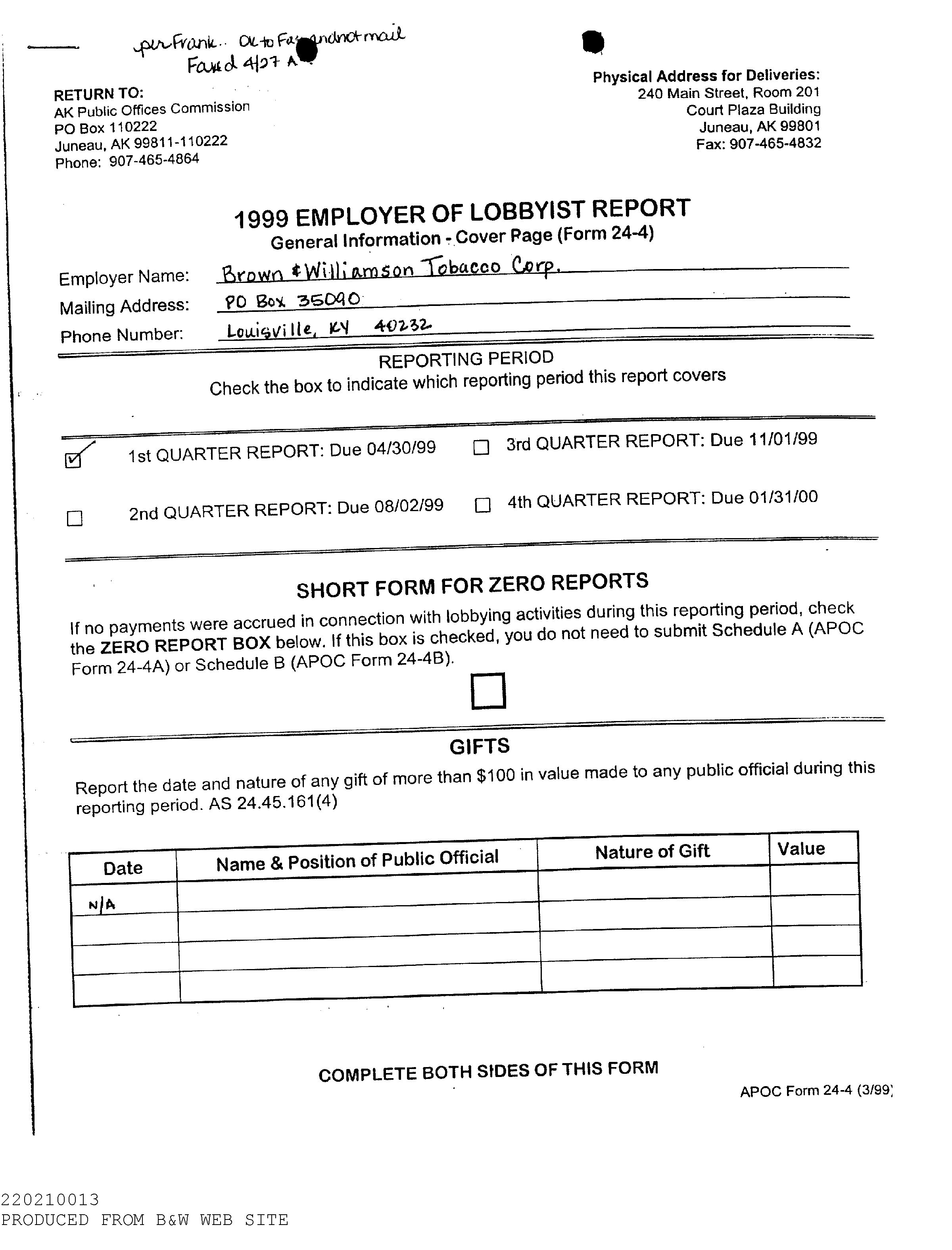}}\hfill
\subfloat{\includegraphics[width=0.40\linewidth]{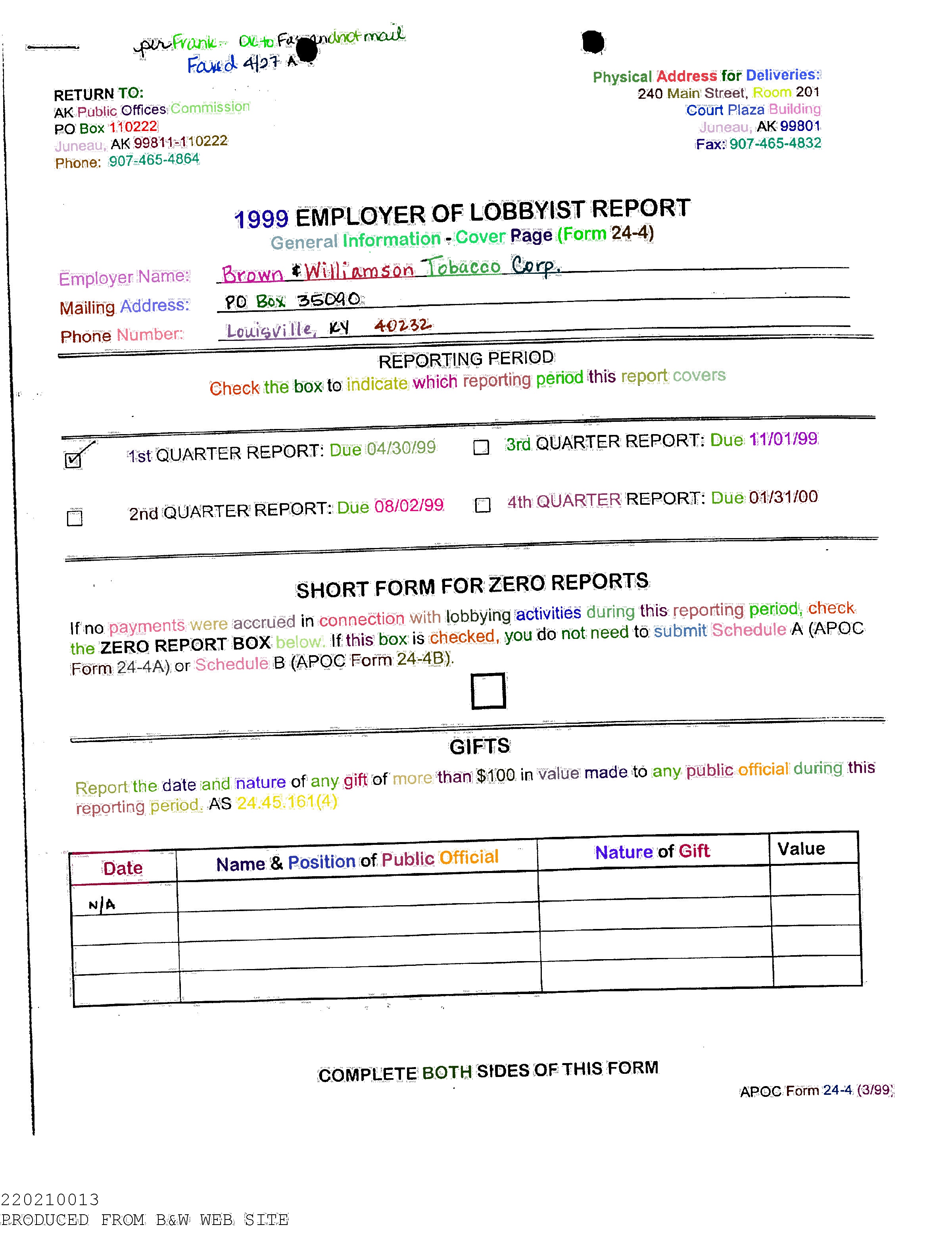}}\hfill
\\
% \subfloat{\includegraphics[width=0.40\linewidth]{wordvisimages/letter.jpg}}\hfill
% \subfloat{\includegraphics[width=0.40\linewidth]{wordvisimages/letter_colorized_cross.jpg}}\hfill
\caption{WordVIS samples of different classes produced. We can see that most of the textual data is masked with colors without changing non-textual elements of the document images.}
\label{fig:samples}
\end{figure}

\subsection{Score Assignment}
Considering all the characters $N_c$ in the language characters space, the limit on supported characters is $\lim_{1\to\infty} N_c$. However, each character in the character space will be assigned a score $C_s$ representing the weightage of that character. The limit for the weightage of a character can be defined as $\lim_{0\to255} C_s$, as the maximum score possible for the $\sum_{n=0}^{n=\infty} C_s$ is $255$ per channel as that's the maximum value possible in the individual channels of the RGB representation of the color. Given the target dataset is Tobacco-3482, all references in this research publication will point to the use of English language characters. However, it is to be noted that the method itself is language agnostic.

\subsubsection{Purpose}
The score assignment process is about assigning weightage $C_s$ to individual characters. This allows the users of the system to leverage weights on individual characters depending on their semantic understanding of their dataset and the need for it. This weightage score forms the basis for the RGB number of the word. If the weights are well distributed on the maximum aforementioned range would give us a unique color for all possible words, as the maximum possible combinations using this scheme are $16777216$ which is a far greater number than all possible words in the English language.

\subsubsection{Distribution Scheme}
The method is flexible to the color score distribution scheme adopted by the users in accordance with the desired output color range. As the number of maximum possible scores per individual character is $NC_s = 3$ one each for the R,G and B channels, the coloring scheme can be based on a single character having a score in all three channels, two out of three or a single channel. 

\subsubsection{Experiment Scores}
For our experiments, there are $26$ English language characters and $10$ digits taking our $N_c = 36$. The assignment operation can be both generic and very targeted towards a specific dataset leveraging the semantic understanding of that dataset. However, for a more fair comparison of the standard dataset, we have chosen to assign more generic scores to the characters of the English language without the use of special characters or semantic understanding of the test dataset. Our range of weights for our experiments is $\lim_{1\to9} C_s$  however, we believe they can be used to further enhance results with little insights into the dataset. Our experimentation scoring scheme is the division of all available characters between Red, Green, and Blue channels with each channel having unique 12 characters. Our score assignment per character is generic in ascending order for all characters and without the use of special characters for avoiding any semantic encoding using the dataset knowledge.

% For reference, our scoring is dsitri of 26 letters in English alphabets were split into three subsets. with the Red and Green channel each getting 9 characters and the remaining 8 characters into Blue channel. Similarly, the number characters were split into Red and Green receiving 3 characters and Blue channel receiving 4.
% \begin{figure}[tbp]
%     \centering   
%         \includegraphics[width=.5\textwidth]{wordvisimages/scoring.png}
%     \caption{color scores of individual channels used in our experiments}
%     \label{fig:coloring}
% \end{figure}

\begin{figure}[t!]
\centering
\captionsetup[subfigure]{labelformat=empty}
\captionsetup[subfloat]{farskip=2pt,captionskip=2pt,singlelinecheck=false,justification=centering}
\subfloat{\includegraphics[width=0.45\linewidth]{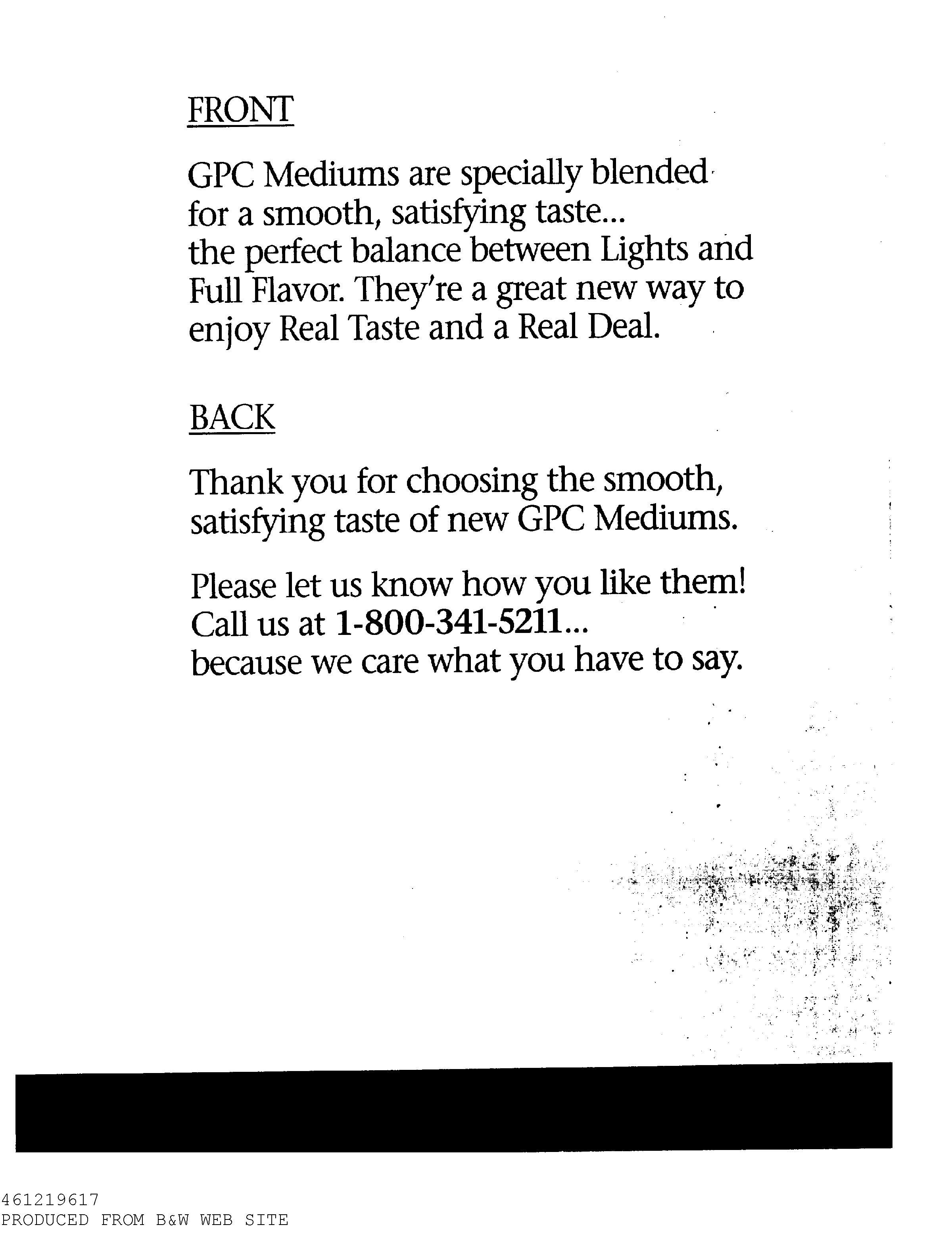}}\hfill
\subfloat{\includegraphics[width=0.45\linewidth]{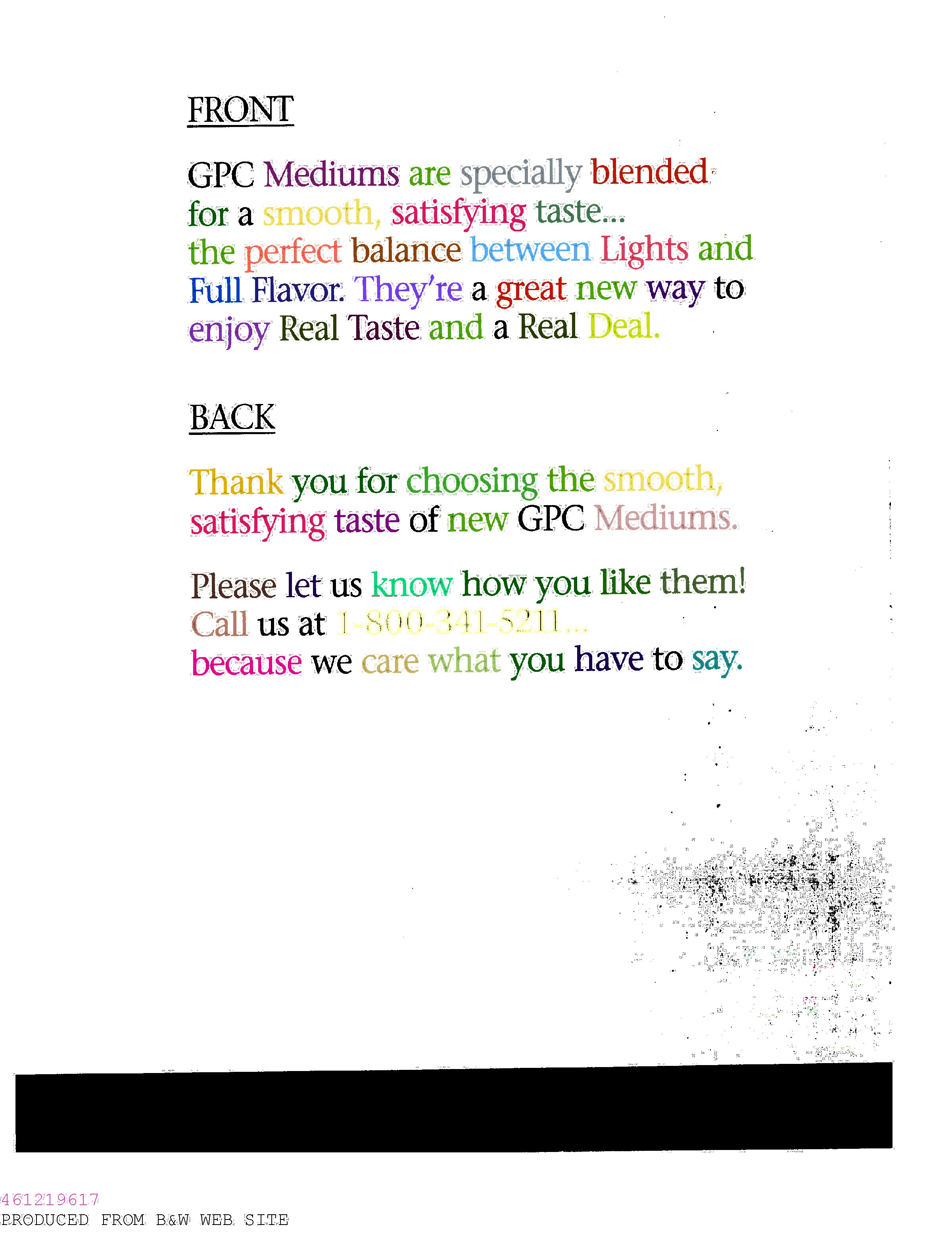}}\hfill
\caption{In WordVIS colorized document we can see that the colorization adapts a pattern of stop words or words more similar to stop words adapting a green color, whereas the more lengthier and distinct words adapting more distinct colors.}
\label{fig:comparison}
\end{figure}

\subsection{Multiplying Factor M}
In order to bring a weight on the length of the word that occurs in the text corpus, we introduced a multiplying factor $M_f$ based on the length of the word. The Multiplying Factor is what ensures that minimally assigned scores in the character space are elevated enough to translate into sharp colors based on the word length. M can be both heuristically assigned according to insights into the data as well as set more generically as we did with our multiplying factor being derived from the word character length itself. In our case, every word will have this factor differently assigned from its character length, and hence why lengthier words will translate to stronger colors than shorter words. This value in combination with the $C_s$ can be used to build a tighter range of the $\sum C_s$.

\subsection{Calculating RGB Values}
The formula that is used for converting the individual scores to RGB is given below in Fig. \ref{eq:wordvis} and Fig. \ref{eq:wordvis2}

\begin{equation}
    R = \sum_{C=a}^{C=i} C_s * M_f, 
    G = \sum_{C=j}^{C=r} C_s * M_f, 
    B = \sum_{C=s}^{C=z} C_s * M_f
    \label{eq:wordvis}
\end{equation}

\begin{equation}
    RGB_\text{color} = (R, G, B)
    \label{eq:wordvis2}
\end{equation}

Whereas $C_s$ is the score of the individual character and $M_f$ is the multiplying factor derived from $Word_\text{length}$.

\subsubsection{Example}
The above algorithm can be more fairly understood from the following calculation example:
Given the word "deep", the RGB score according to the above example would be \\

$Word_\text{length}$ = 4 = M \\
$R_\text{score}$ = ((d=3) * 4) + ((e=5) * 4) + ((e=5) * 4) = 52 \\
$G_\text{score}$ = ((p=7) * 4) = 28 \\
$B_\text{score}$ = ((0) * 4) = 0 \\

$RGB_\text{score}$ = $(52, 28, 0)$\\

This example highlights a property of WordVIS, where small errors in OCR are limited by the constraints put on the scoring mechanism leading to more consistent colors even in case of errors, to further continue the previous example:
If the given word "deep" was incorrectly OCR'ed as "deeq", the resulting change would be minimal as shown in the calculation below.
$Word_\text{length} = 4 = M$\newline
$R_\text{score} = ((d=3) * 4) + ((e=5) * 4) + ((e=5) * 4) = 52$\newline
$G_\text{score} = ((p=8) * 4) = 32$\newline
$B_\text{score} = ((0) * 4) = 0$\newline

$RGB_\text{color} = (52, 32, 0)$\newline

\subsection{Output and Sample Analysis}
The following samples shows the resulting documents of several classes from the WordVIS visual embedding on Tobacco-3482 documents:

% \begin{figure}[tbp]
%     \centering   
%         \includegraphics[width=1\textwidth]{wordvisimages/samples.png}
%     \caption{Colored Documents}
%     \label{fig:sample1}
% \end{figure}

Fig. \ref{fig:samples} shows that the images are pre-processed utilizing varying degrees of colors using the scoring mechanism. Through further inspection, we notice that most of the stop words due to them containing the same letters repeating, contain the same colors and non-stop words tend to take more different colors than the stop words. This property can be observed in all classes, from our Fig. \ref{fig:comparison} in class ADVE it can be observed that words like "you, are, the, and, for, new, what" all have green hues, whereas words such as "satisfying, blended, taste, medium, smooth" all take sharper and different tones of colors.

\section{Experiments and Results}
To evaluate the efficacy of our proposed method we trained several architectures on the standard dataset of Tobacco-3482. The smaller dataset was used in order to test for the performance gain on small limited amounts of datasets. We trained all these architectures on two different versions of the dataset.

\begin{itemize}
    \item Standard Tobaccoo-3482 dataset
    \item WordVIS Colorized Tobaccoo-3482 dataset
\end{itemize}

\begin{table*}[!t]
	\centering
	\caption{A comparison of the classification accuracy of different approaches on the Tobacco3482 datasets}
	\begin{center}
		\begin{adjustbox}{max width=\textwidth}
			\begin{tabular}{llcccccc} 
				\toprule
			     Model&\parbox{3cm}{\centering Inference\\Time [ms]}&\# of Parameters&\parbox{3cm}{\centering Tobacco3482\\(ImageNet pre-training)}\\
				\midrule
				
				AlexNet (Afzal \etal, 2017 \cite{errorbyhalf})&\parbox{3cm}{\centering 1.1}&57M&75.73\%\\
				\CC GoogleNet (Afzal \etal, 2017 \cite{errorbyhalf})&\CC \parbox{3cm}{\centering 1.2}&\CC5.6M&\CC 72.98\%\CC \\
				ResNet-50 (Afzal \etal, 2017 \cite{errorbyhalf})&\parbox{3cm}{\centering 1.1}&23.5M&67.93\%\\
				\CC VGG-16 (Afzal \etal, 2017 \cite{errorbyhalf})&\CC\parbox{3cm}{\centering 1.3}&\CC134M&\CC 77.52\%\CC \\
				EfficientNet (Ferrando \etal, 2020 \cite{doc-img-cls-effnet-Ferrando2020})&\parbox{3cm}{\centering 2.3}&17.6M& 85.99\%\\
				\CC EfficientNet + BERT (Ferrando \etal, 2020 \cite{doc-img-cls-effnet-Ferrando2020})&\CC\parbox{3cm}{\centering -}&\CC 127.6M&\CC 89.47\%\\
                MobileNetV2+Text (Audebert \etal, 2019 \cite{audebert})&\parbox{3cm}{\centering -}&-&87.80\%\\
				\CC EfficientNet+BERT (Kanchi \etal, 2022 \cite{kanchi})&\CC\parbox{3cm}{\centering -}&\CC 197M&\CC 90.3\%\\
                \CC DocXClassifier-B/384 &\CC \parbox{3cm}{\centering 6.53}&\CC 95.4M&\CC 88.42\%\\
				DocXClassifier-L/384 &\parbox{3cm}{\centering 10.0}&204M&88.43\%\\
				\CC DocXClassifier-XL/384 &\CC \parbox{3cm}{\centering 16.1}&\CC 356M&\CC 90.14\%\\
                \textbf{WordVIS DocXClassifier-B/384 (OURS)} &\parbox{3cm}{\centering \textbf{6.53}}&\textbf{95.4M}&\textbf{91.14\%}\\
				\bottomrule
			\end{tabular}
	   \end{adjustbox}
    \end{center}
	\label{table:accuracy}
\end{table*}
\subsection{Dataset}
The dataset consists of a total of 3482 documents split across 10 classes. The splits were generated using random $80\% - 20\%$ train-test split resulting in $700$ test documents. The validation split was $10\%$ of the training data resulting in $2504$, $278$, $700$ samples in Train, Validation, and Test respectively. Moreover, it is to be noted that these experiments were performed multiple times with the given splits to eliminate the standard deviation as the cause for higher accuracy.

\subsection{Training Details}
For hyperparameters consistency, we ensure the same hyperparameters for both with and without the use of WordVIS to eliminate any issues that could be caused by differences in hyper parameters. For training the DocXClassifier-B the original hyperparameters used by authors of DocXClassifier Saifullah et al \cite{doc-img-cls-docxclassifier-saifullah2022} were used. All the remaining architectures of ResNet50, ResNet101, densenet121, EfficientNetV2 were trained with the same hyperparameters. For the optimizer Stochastic Gradient Decent (SGD) was used with a learning rate of: 0.5, weight decay was set to 1.0e-8, and batch size of 64. All the architectures were trained for 300 epochs, however, the convergence occurred for all before the 80 epoch mark. 
ferrandodc
\subsection{Evaluation}
For comparative analysis using a state-of-the-art document classifier on Tobacco-3482 dataset, we took the DocXClassifier-B, the smallest in terms of the number of parameters in the state-of-the-art category, and tested WordVIS. As seen in Table \ref{table:accuracy}, our method WordVIS improves on the base model drastically, setting a new record score on the Tabaccoo-3482 dataset of $91.14$\%. It is to be noted that not only does it improve drastically on the base class but also outperforms the previous state-of-the-art DocXClassifier-XL while using $73.2 \%$ fewer parameters and reducing the inference time by $60\%$.

\begin{figure}[H]
\centering
\captionsetup[subfigure]{labelformat=empty}
\captionsetup[subfloat]{farskip=2pt,captionskip=2pt,singlelinecheck=false,justification=centering}
\subfloat[ADVE-WordVis]{\includegraphics[width=0.40\linewidth]{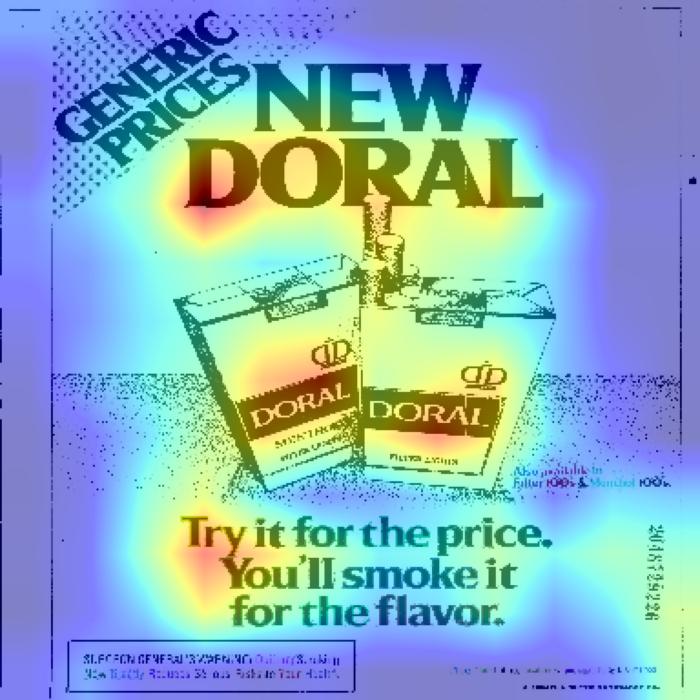}}\hfill
\subfloat[ADVE-w/o WordVis]{\includegraphics[width=0.40\linewidth]{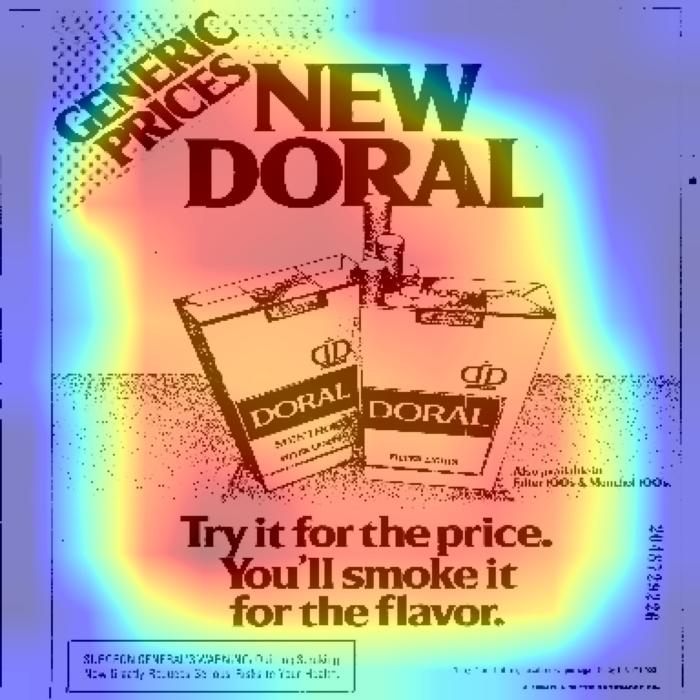}}\hfill
\\
\subfloat[News-WordVis]{\includegraphics[width=0.40\linewidth]{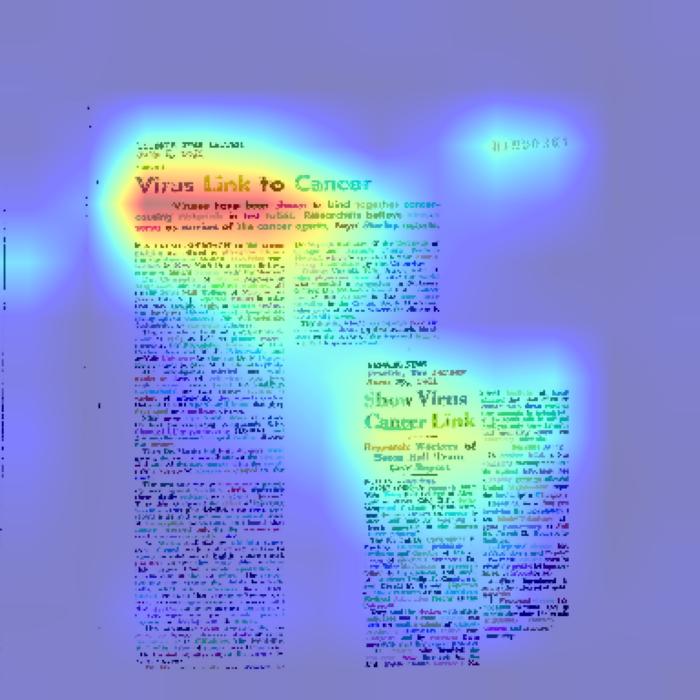}}\hfill
\subfloat[News-w/o WordVis]{\includegraphics[width=0.40\linewidth]{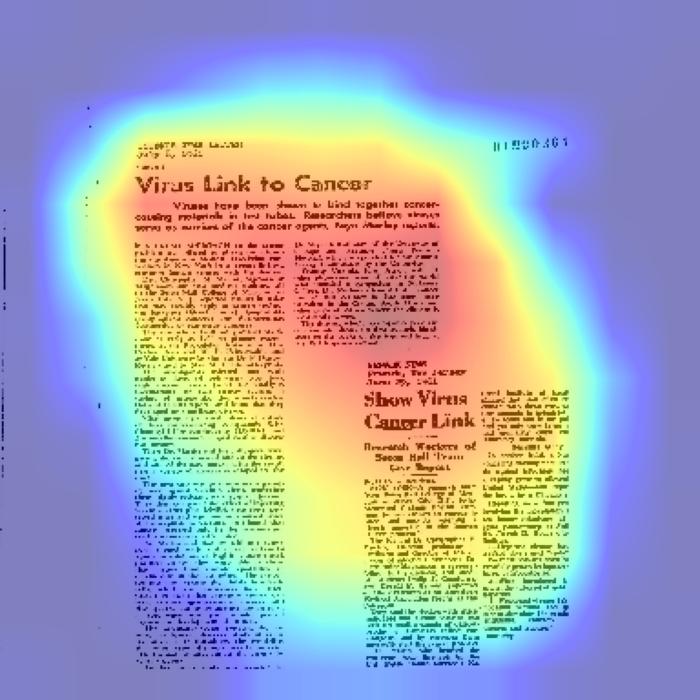}}\hfill
\\
\subfloat[Email-WordVis]{\includegraphics[width=0.40\linewidth]{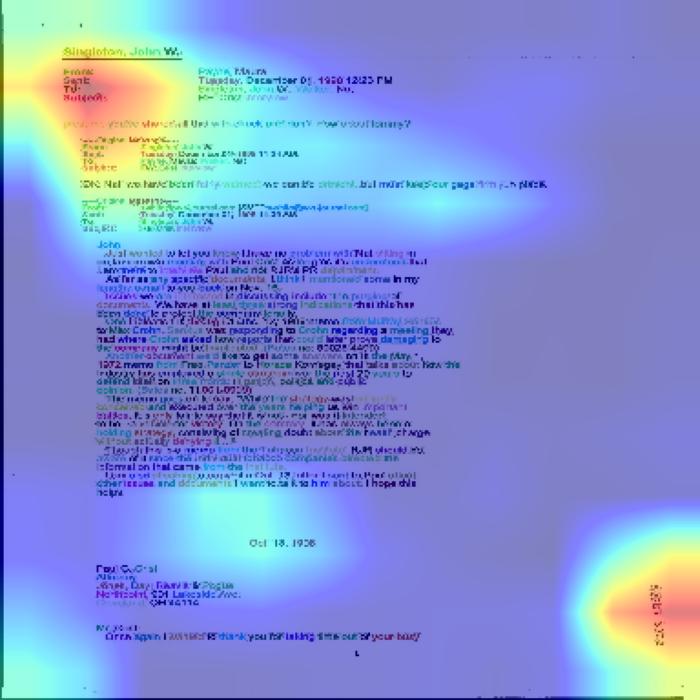}}\hfill
\subfloat[Email-w/o WordVis]{\includegraphics[width=0.40\linewidth]{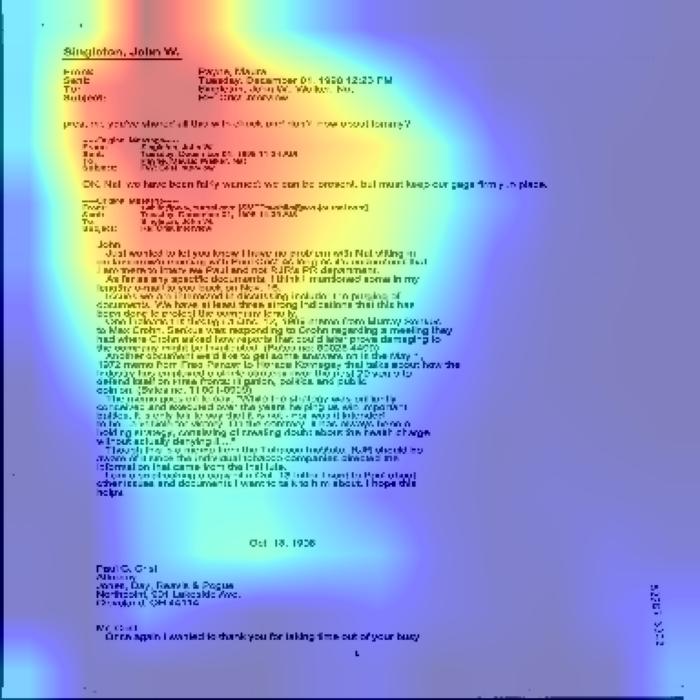}}\hfill
\\
\subfloat[Form-WordVis]{\includegraphics[width=0.40\linewidth]{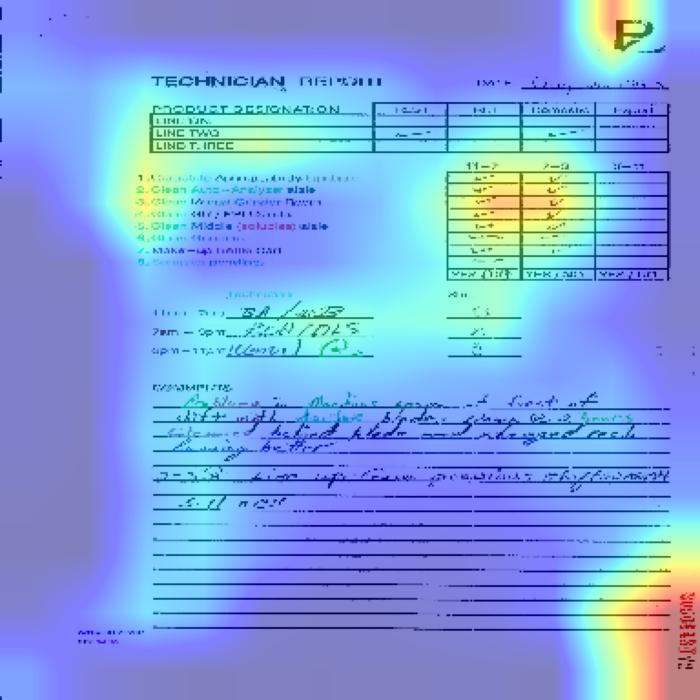}}\hfill
\subfloat[Form-w/o WordVis]{\includegraphics[width=0.40\linewidth]{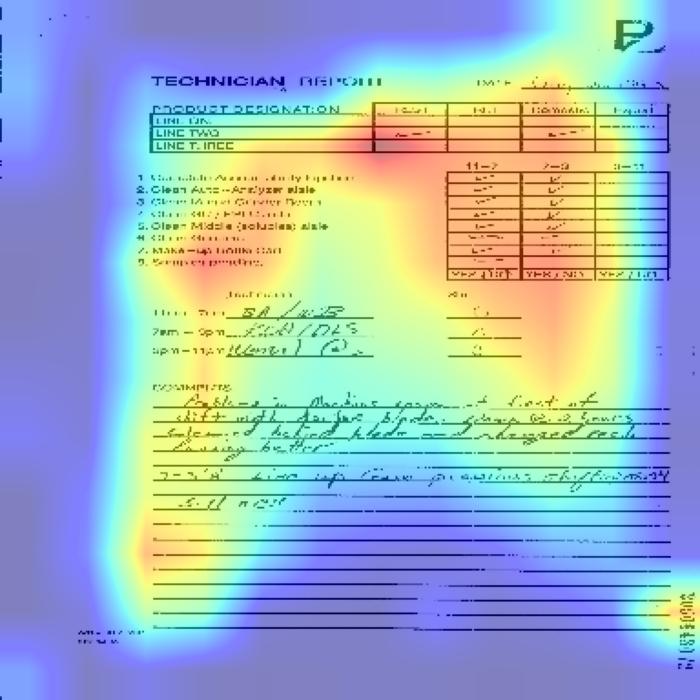}}\hfill

\caption{Sample heatmaps generated using DocXClassifier-B with WordVIS (Left) and Without WordVIS (Right)}
\label{fig:heatmaps}
\end{figure}

\subsection{Ablation Study}

To further test our method, we also trained several CNN variant architectures. As we can see from Table \ref{table:accuracy2} that the WordVIS method outperformed the no pre-processing on all architectures.

\begin{table*}[b]
	\centering
	\caption{Ablation study: A comparison of different architectures with and without WordVIS}
	\begin{center}
		\begin{adjustbox}{max width=\textwidth}
			\begin{tabular}{llcccccc} 
				\toprule
				Model&\parbox{6cm}{\centering Base}&\parbox{3cm}{\centering WordVIS\\}\\
				\midrule
                ResNet50 &\parbox{6cm}{\centering$67.9$\%}&\textbf{72.5\%}\\
				ResNet101 &\parbox{6cm}{\centering$79.1$\%}&\textbf{82.0\%}\\
                densenet121 &\parbox{6cm}{\centering$77.6$\%}&\textbf{80.3\%}\\
                EfficientNetV2 &\parbox{6cm}{\centering$74.1$\%}&\textbf{76.3\%}\\
				\bottomrule
			\end{tabular}
	   \end{adjustbox}
    \end{center}
	\label{table:accuracy2}
\end{table*}

\begin{figure}[!t]
\centering
\captionsetup[subfigure]{labelformat=empty}
\captionsetup[subfloat]{farskip=2pt,captionskip=2pt,singlelinecheck=false,justification=centering}
\subfloat{\includegraphics[width=0.45\linewidth]{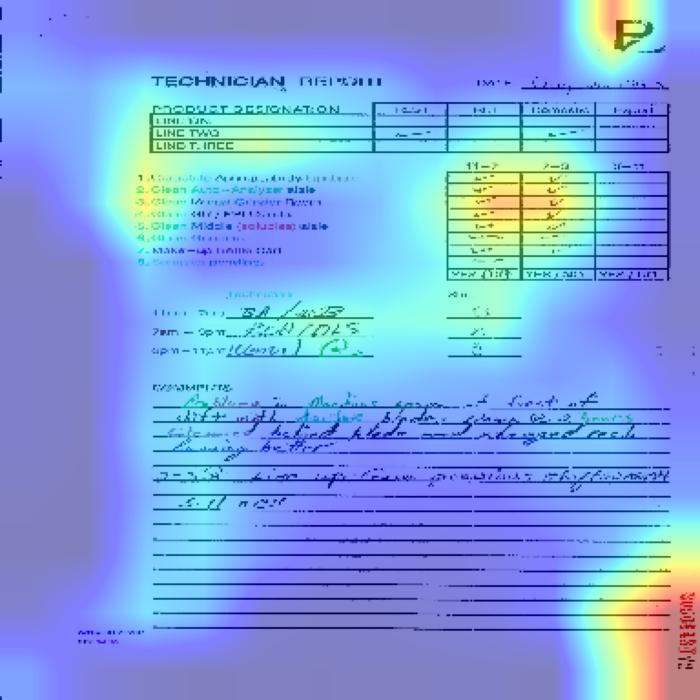}}\hfill
\subfloat{\includegraphics[width=0.45\linewidth]{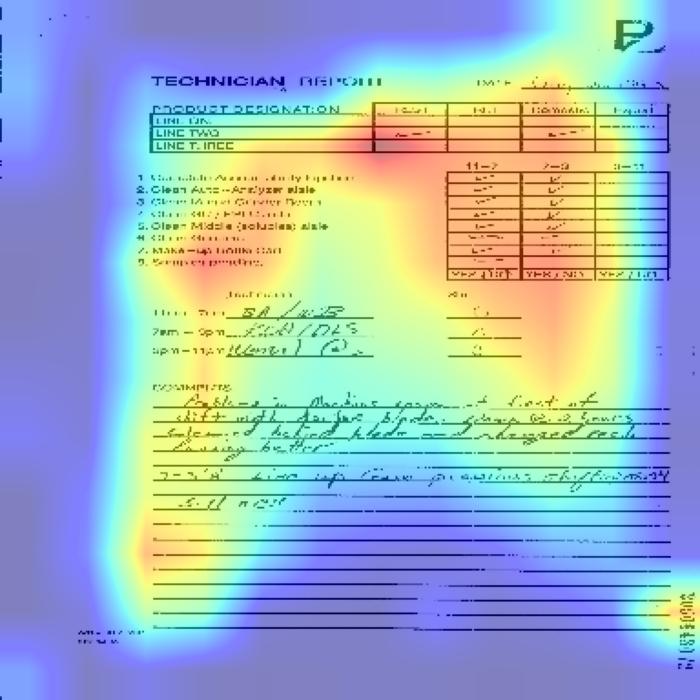}}\hfill
\caption{Heatmaps Form: Form heatmaps also gives us clues into how the WordVIS (Left) trained network focuses on boxes and the content inside the boxes as opposed to the Base (Right).}
\label{fig:heatmaps_form}
\end{figure}

\begin{figure}[t!]
\centering
\captionsetup[subfigure]{labelformat=empty}
\captionsetup[subfloat]{farskip=2pt,captionskip=2pt,singlelinecheck=false,justification=centering}
\subfloat{\includegraphics[width=0.45\linewidth]{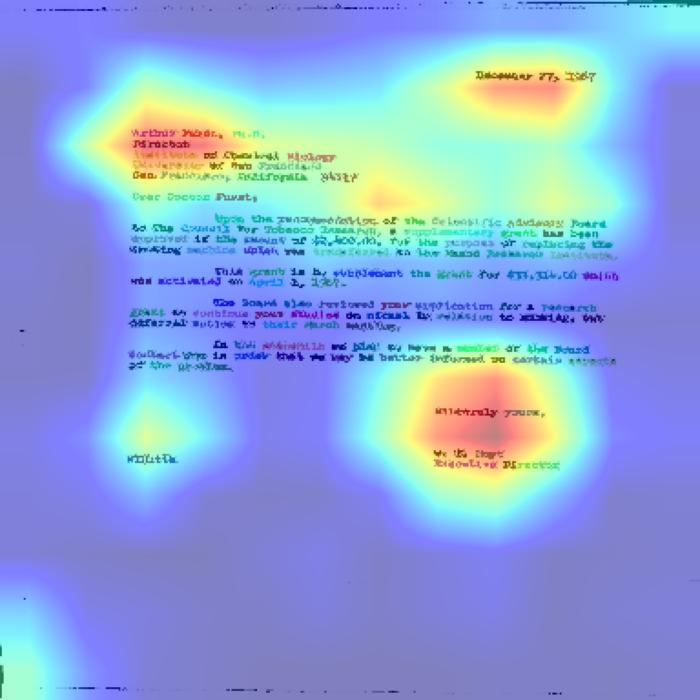}}\hfill
\subfloat{\includegraphics[width=0.45\linewidth]{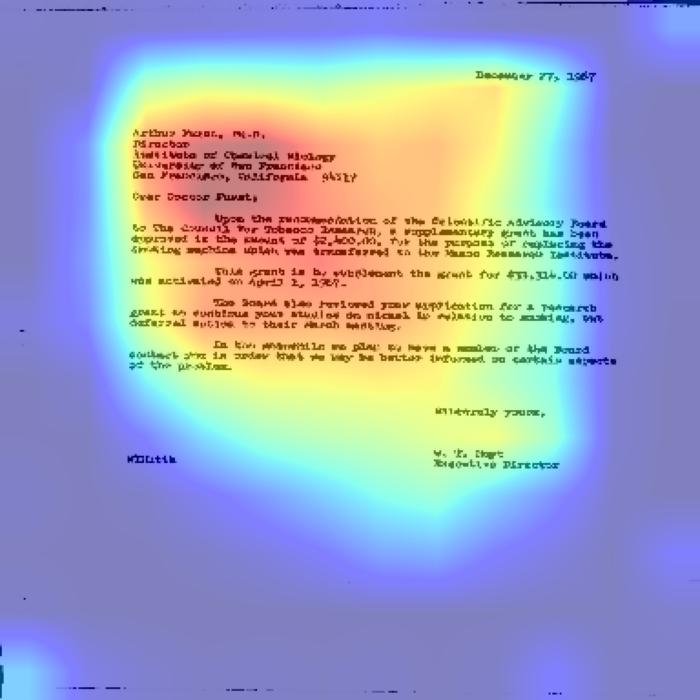}}\hfill
\caption{Heatmaps of Letters shows us more more in depth on how the WordVIS method (Left) helps the network focus in on specific textual elements unique to the letter, whereas the training without the WordVIS focuses on the image as a whole}
\label{fig:letter}
\end{figure}

\subsection{Qualitative Analysis}
In order to analyze and verify the improvements brought in by WordVIS, we generated activation HeatMaps for test samples. As seen from Fig. \ref{fig:heatmaps} we can see that the WordVIS (Top Row) shows a more refined focus toward specific text areas instead of a generalized focus on the entire image space, suggesting that the network learns to follow the text. In the example of Form Fig. \ref{fig:heatmaps_form} we can see that the WordVIS-trained DocXClassifier-B focuses on the specific boxed area and the activation is quite well tightly bounded as opposed to the base-trained DocXClassifier-B. Additionally, going through samples of Letters as depicted in Fig, \ref{fig:letter} we observe a pattern of the WordVIS-trained network focusing on letter headers and enclosers specifically rather than the entire body, in essence, closer to how humans identify letters as opposed to the Base trained network where the activation is spread on the entire image.

% \begin{figure}[t!]
% \centering
% \captionsetup[subfigure]{labelformat=empty}
% \captionsetup[subfloat]{farskip=2pt,captionskip=2pt,singlelinecheck=false,justification=centering}
% \subfloat{\includegraphics[width=0.24\linewidth]{heatmaps/ADVE-2048729226_colorized.jpg}}\hfill
% \subfloat{\includegraphics[width=0.24\linewidth]{heatmaps/News-10030325_colorized.jpg}}\hfill
% \subfloat{\includegraphics[width=0.24\linewidth]{heatmaps/Email-522873332_colorized.jpg}}\hfill
% \subfloat{\includegraphics[width=0.24\linewidth]{heatmaps/Form-2050648072_colorized.jpg}}\hfill
% \\

% \subfloat{\includegraphics[width=0.24\linewidth]{heatmaps/ADVE-2048729226.jpg}}\hfill
% \subfloat{\includegraphics[width=0.24\linewidth]{heatmaps/News-10030325.jpg}}\hfill
% \subfloat{\includegraphics[width=0.24\linewidth]{heatmaps/Email-522873332.jpg}}\hfill
% \subfloat{\includegraphics[width=0.24\linewidth]{heatmaps/Form-2050648072.jpg}}\hfill

% \caption{Sample Heatmaps Generated using DocXClassifier-B with WordVIS (Top Row) and Base (Bottom Row)}
% \label{fig:heatmaps}
% \end{figure}

\section{Conclusion and Future Work}
In this research publication, we presented WordVIS, a novel pre-processing approach for utilizing text in Document Classification on smaller datasets. The heatmap analysis conducted in this research proves that our approach successfully embedded textual data in visual space, leading to the elimination of the textual embedding overhead used in multi-modal approaches. The research also successfully proves that image-based networks can be used to learn contextual text data by enhancing the textual data in document images. This approach not only reduces the required training data and compute overhead but also is able to efficiently leverage small amounts of data to outperform larger networks. Furthermore, the heatmap analysis also proves that such textual coloring techniques not only improve the quantitative accuracy but also drastically improve the quantitative results of the document classifiers. Our goal with this research was to enable businesses with limited training data and compute resources to be able to use and leverage document classifiers for their business use cases. We feel this is a step in the right direction, however, more research needs to be conducted to expand our work in order to improve accuracy on relatively simpler Deep Learning networks and reduce the architectural complexity requirements for more tolerable results.     

\bibliographystyle{splncs04}
\bibliography{main}
\end{document}